\pdfoutput=1

\documentclass[11pt]{article}

\usepackage[preprint]{acl}
\usepackage{times}
\usepackage{latexsym}

\usepackage[T1]{fontenc}

\usepackage[utf8]{inputenc}

\usepackage{microtype}

\usepackage{inconsolata}

\usepackage{microtype}
\usepackage{graphicx}
\usepackage{multirow}
\usepackage{amsmath}
\usepackage{caption}
\usepackage{xcolor}

\usepackage{latexsym}
\usepackage{subcaption}

\usepackage{color}
\usepackage{array}
\usepackage{booktabs} 
\usepackage{times}

\usepackage{hyperref}
\DeclareMathOperator*{\argmax}{arg\,max}

\title{Inconsistencies in Masked Language Models}

\author{Tom Young \hspace{1cm} Yunan Chen \hspace{1cm} Yang You \\               
        School of Computing, National University of Singapore, Singapore \\
        \texttt{tomyoung903@gmail.com}, \texttt{chen.yunan\_01@u.nus.edu}, \texttt{youy@comp.nus.edu.sg} \\
        Code: \url{https://github.com/tomyoung903/MLM_inconsistencies/tree/master}
        }

\begin{document}
\maketitle
\begin{abstract}
Learning to predict masked tokens in a sequence has been shown to be a helpful pretraining objective for powerful language models such as PaLM2. After training, such masked language models (MLMs) can provide distributions of tokens in the masked positions in a sequence. However, this paper shows that distributions corresponding to different masking patterns can demonstrate considerable inconsistencies, i.e., they cannot be derived from a coherent joint distribution when considered together. 

This fundamental flaw in MLMs can lead to self-contradictory behaviors during inference. On various benchmark datasets including MMLU, MLMs can give different predictions to the same input question. From BERT-base to UL2-20B, we show that such inconsistencies exist ubiquitously in MLMs of diverse sizes and configurations. In light of our observations, we further propose an inference-time strategy for MLMs called Ensemble of Conditionals. It jointly considers a selected range of inconsistent conditionals directly produced by the MLM for the final prediction, which often leads to considerable accuracy improvement.
\end{abstract}

\section{Introduction}

Pretraining objectives of large language models can be roughly divided into two categories. First, vanilla next token prediction (also known as casual language modeling) aims to learn the distribution of the next token in a sequence given the context to the left \cite{brown2020language}. Second, the masked language modeling (MLM) objective, which masks out a portion of the tokens in a sequence and asks the model to predict them, aims to learn the distribution of one or more tokens given surrounding context \cite{devlin2018bert, raffel2020exploring}.

While GPT-3 \cite{brown2020language} used vanilla next token prediction, following work such as PaLM-2 \cite{anil2023palm}, U-PaLM \cite{tay2022transcending}, GPT-FIM \cite{bavarian2022efficient}, UL2 \cite{tay2022unifying}, and GLM \cite{zeng2022glm} have hinted that incorporating the MLM objective could be highly beneficial to performance. In addition, \citet{tay2022transcending} has demonstrated that such bidirectional conditionals provide strong infilling capabilities. Empirically speaking, predicting masked tokens in the middle of the sentence can be seen as a natural data augmentation technique to vanilla next token prediction, which might be helpful to alleviating the data scarcity problem \cite{xue2023repeat} in the current large model era.

One may notice that, unlike the unidirectional conditional distributions that vanilla next token prediction learns, the bidirectional conditionals that MLMs learn are overly abundant in terms of representing a coherent joint distribution. Therefore, they are not guaranteed to be self-consistent. This paper explains our effort on exposing and quantifying this issue and corresponding strategies during inference.

\begin{figure}[h]
    \centering
    \includegraphics[width=1\linewidth]{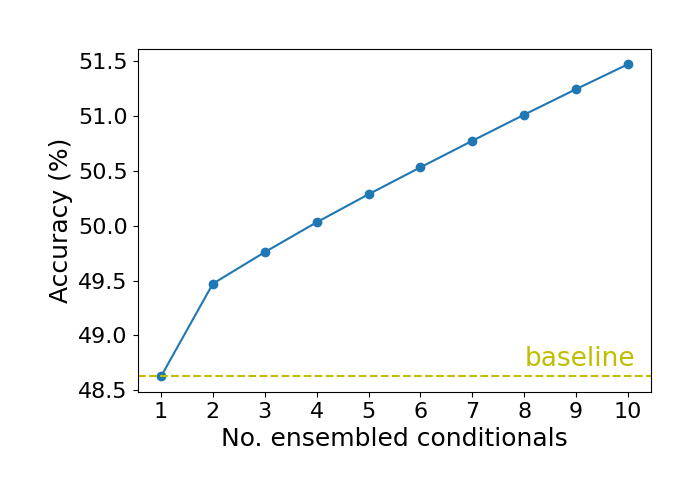}
    \caption{Self-ensembling improves MLMs' accuracies on standard benchmarks including MMLU, Lambada and BigBench. Aggregated results based on Figure \ref{fig:eoc_accuracy}.}
    \label{fig:eoc_accuracy_aggregate}
\end{figure}

\begin{figure*}[h]
\centering
 \includegraphics[width=1.0\textwidth]{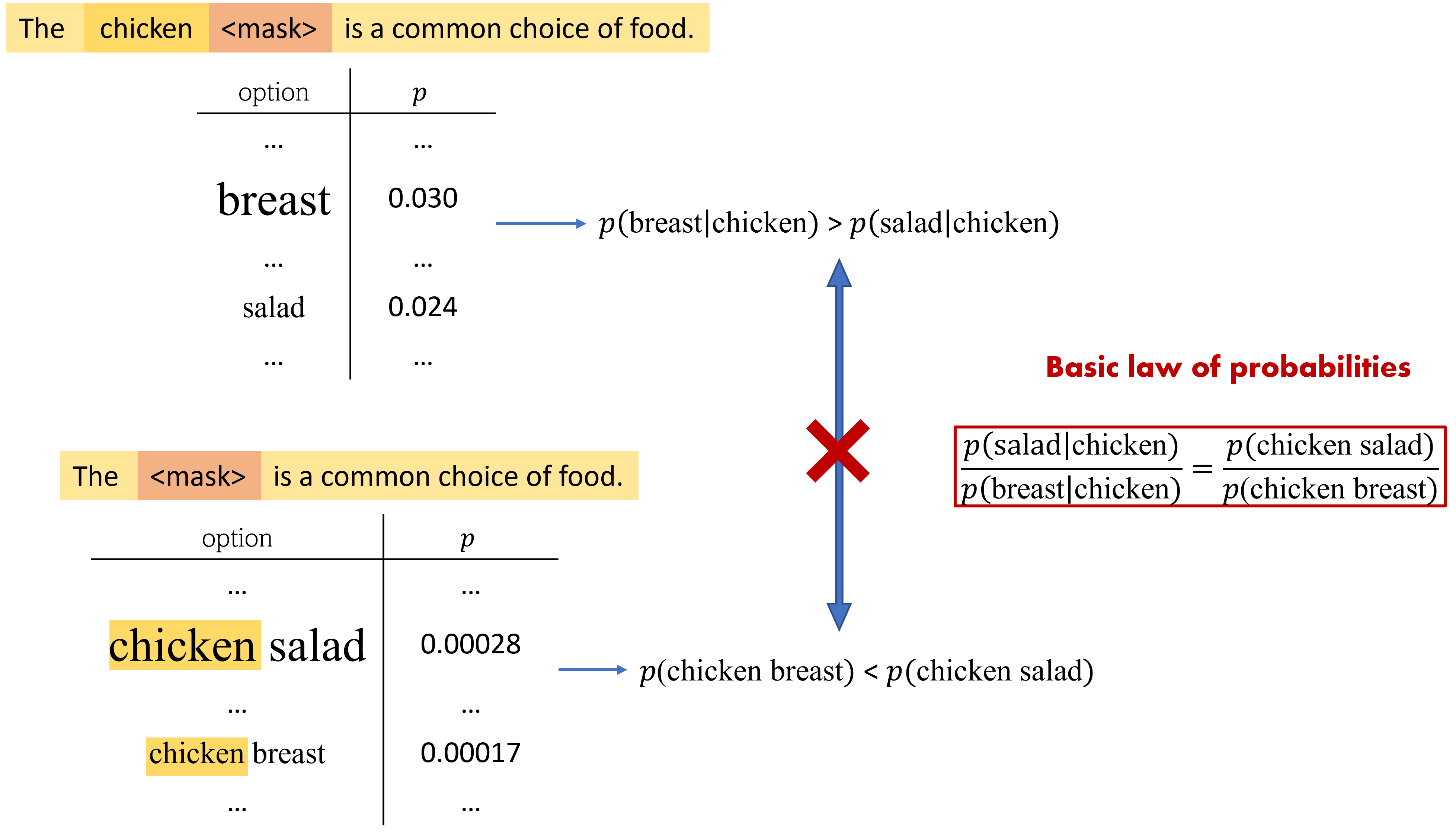}
\caption{A simple bigram comparison example that exposes the inconsistencies in the T5 model. The conditional probabilities that the model learned (quoted from T5-11B fed with the shown masked sequences) contradict each other greatly. Not only are the ratios unbalanced, the model confuses its own preference of the two bigrams.}\label{fig:t5_example}
\end{figure*}

To begin with, a simple example for such inconsistencies is shown in Figure \ref{fig:t5_example}. In this example, we obtain the bidirectional conditional distributions that the T5 model learned using two input masked sequences. The two similar sequences are designed with a small difference, in order to examine if the resulting conditionals satisfy a basic law of probabilities (hold consistency). Results clearly show otherwise. We design experiments to quantify such inconsistencies on benchmark datasets in Section \ref{sec:exposing}.
We further show an inference-time ensemble algorithm in Section \ref{sec:ensemble} which utilizes many inconsistent conditionals for a more accurate prediction. We demonstrate that ensembling the numerous inconsistent conditionals directly provided by the MLM can improve its performance (Figure \ref{fig:eoc_accuracy_aggregate}).

In summary, our contributions are
(1) We expose the commonly overlooked flaw in MLMs that they can represent inconsistent distributions depending on the mask patterns.
(2) We quantify such inconsistencies in benchmark datasets including Lambada \cite{paperno2016Lambada}, MMLU\cite{hendrycks2021measuring} and BigBench \cite{srivastava2023beyond}. For example, on multiple choice questions in MMLU, 2 different distributions given by UL2-20B disagree on the answer 14\% of the time on average.
(3) We show that the numerous inconsistent conditionals can be ensembled together to considerably improve accuracy on said benchmarks.

\section{Why inconsistencies can occur in MLMs}\label{sec:why}

For a set of conditional distributions to be self-consistent, they need to be able to be derived from a single coherent joint distribution. 

One essential reason for the inconsistencies to occur among the conditionals provided by a trained MLM is that the number of conditionals it can provide far \textit{exceeds} the degrees of freedom of a joint distribution.

Consider a sequence of length $L$ with vocabulary $V$. The joint distribution of the tokens in such a sequence is defined by $|V|^L$ probabilities that sum to 1. Therefore, the degrees of freedom ($D$) of such a joint distribution is:
\begin{align}
D_{joint}=|V|^L - 1,
\label{eq1}
\end{align}

Both vanilla next token prediction models and MLMs essentially learn conditionals that predict some tokens in the sequence given others. Such conditional probabilities and probabilities from the joint distribution can be linearly derived from each other. Therefore, each free conditional that the language model is capable of specifying places a constraint on the joint distribution. One can easily verify (by counting the conditionals left to right for a geometric sequence) that a vanilla next token prediction based language model provides just $|V|^L - 1$ free conditionals\footnote{A single softmax operation over $V$ essentially gives $|V| - 1$ free conditionals. Here we call conditionals free when they can be assigned any values decided by an underlying neural network.} to exactly determine the joint distribution. Therefore, a vanilla next token prediction model (no matter how it is trained, or even untrained) would never suffer from inconsistencies among its conditionals.

MLMs, which can provide distributions of masked tokens given bidirectional context, could specify far more free conditionals. For the simplest case, where the MLM predicts the distribution of only 1 (masked) token given $L-1$ other (unmasked) tokens in the sequence,  the total number of free conditionals ($N$) is 
\begin{align}
N_{mlm}(1)= L \times (|V|^L - |V|^{L-1}),
\label{eq2}
\end{align}

Just $N_{mlm}(1)$ is already far larger than $D_{joint}$. Not to mention $N_{mlm}(k)$ for $k \in [2, N-1]$. See Appendix \ref{sec:no_conditionals} for $N_{mlm}(k)$ and both of their derivations. The fact that the number of conditionals an MLM provides far exceeds what is needed for defining a joint distribution sets up room for there to be inconsistencies among them.

The first portion of our experiments (Sections \ref{sec:exposing} \& \ref{sec:BERT}) focus on exposing and quantifying the inconsistencies that exist among the conditionals provided by common MLMs. The second portion of our experiments (Section \ref{sec:ensemble}) demonstrates our new inference-time algorithm ``Ensemble of Conditionals'' that unites them for more accurate predictions.

To begin with, the next section explains the backbone models that this paper works with.




\section{Backbone MLMs}\label{Section-backbones}

We work with 3 different MLMs in this paper that belong to two different styles, which can be called the T5-style and the BERT-style.

\subsection{T5-style}

For T5-style MLMs, the definition here is that each mask token in the input functions as a placeholder for the prediction of an entire span of tokens of variable length. Below we introduce 2 different T5-style MLMs that we will work with in the experiments. They differ in their architecture design, masking strategies and sizes.

\begin{enumerate}
    \item T5

    The T5 model \cite{raffel2020exploring} uses an Encoder-Decoder architecture. It uses a corruption rate of 15\% and an average span length of 3 tokens. The masked spans can be anywhere in the sequence. We use the largest model T5-11B in the experiments. 
    
    \item UL2-20B
    
    The UL2 model \cite{tay2022unifying} follows T5's architecture design and aims to mix up 3 masking strategies to more comprehensively utilize the pretraining corpus. The MLM objective is also known as the auto-denoising objective, since the masks can be considered as adding noise to the sequence. UL2 calls masking strategies denoisers.
    
    \textbullet\ The R(Regular)-Denoiser mimics T5's masking scheme. 
    
    \textbullet\ The S(Sequential)-Denoiser simply partitions the input sequence into two consecutive sub-sequences and predict the second sub-sequence as the masked sequence. 
    
    \textbullet\ The X(Extreme)-Denoiser is an extreme version of denoising marked by long corrupted spans or high corruption rates. The X-Denoiser is aimed as an interpolation between R- and S-Denoiser. 
    
    \citet{tay2022unifying} showed that such a mixture of masking strategies achieved a superior performance than T5 on many tasks. The 3 different denoisers were differentiated by 3 respective sentinel tokens ([R], [S], [X]) prepended to the sequence. These sentinel tokens are also used during inference to invoke the corresponding behavior from the model. Without losing generality, we restrict ourselves to the X-Denoiser in our experiments due to its superior performance in our pilot trials.

\end{enumerate}

\subsection{BERT-style}
Our definition for BERT-style MLMs, named after BERT \cite{devlin2018bert}, is that the model uses each mask token as the placeholder for the prediction of exactly one real token. We use the better-trained RoBERTa \cite{liu2019roberta} for our experiments as our example for BERT-style MLMs, which shares the same architecture as BERT. While considered somewhat deprecated \cite{tay2022unifying} compared to later MLMs like T5, UL2 and PaLM2, BERTs are unique in terms of their architecture design because they use a single transformer with bi-directional attention (or, an Encoder-only architecture), as opposed to GPTs \cite{radford2018improving, brown2020language}, which use a transformer with uni-directional attention (Decoder-only) or the T5 model (Encoder-Decoder).

Our paper mainly focuses on the inconsistencies in T5-style MLMs since they are most useful in practice (Section \ref{sec:T5}). But we also touch on BERT-style MLMs due to its unique architecture and historical impact (Section \ref{sec:BERT}).

\section{Inconsistencies in T5-style MLMs} \label{sec:T5}

\subsection{Conditionals for various mask patterns}
\label{sec: conditionals}
This section lists a few different types of conditional distributions that a trained T5-style MLM can give depending on the mask pattern. This sets up for the next two Sections (\ref{sec:exposing} and \ref{sec:ensemble} )which discusses their inconsistencies and how to ensemble them on various benchmark datasets.

\begin{figure*}
    \centering
    \includegraphics[width=1\linewidth]{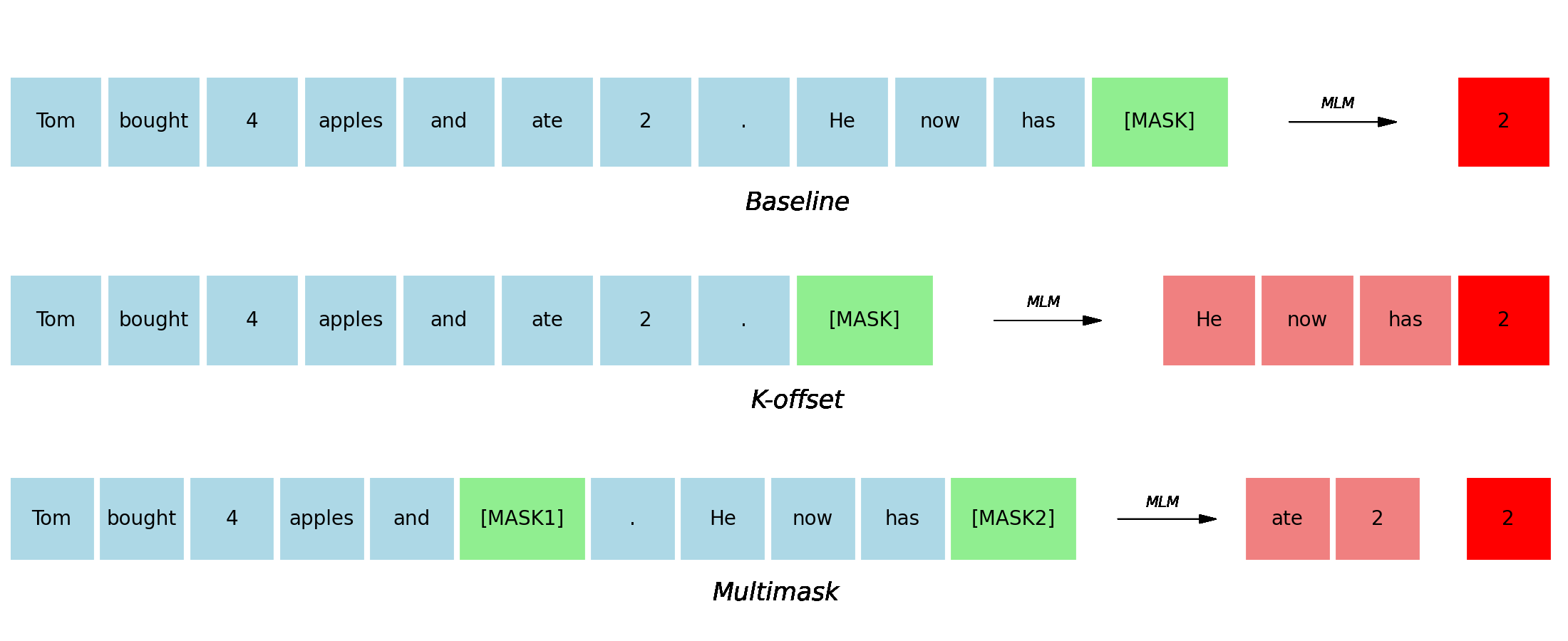}
    \caption{K-offset and Multimask patterns. The goal here is to prompt the MLM for different versions of the target token distribution. The red token is our target token. The coral tokens are taken from the original input sequence and fed as starting tokens to the decoder of the MLM.}
    \label{fig:conditionals}
\end{figure*}

First, we discuss the baseline conditional distribution (first row in Figure \ref{fig:conditionals}). Since most NLP tasks can be formulated as predicting continuing tokens given an input sequence, we consider the use case of MLMs where we append a single \texttt{[MASK]} token behind the input sequence \cite{tay2022unifying}. The MLM takes as input this modified sequence to generate a distribution of tokens for the \texttt{[MASK]} position, which is essentially our distribution of interest for the target tokens.
 
Tweaking the mask pattern can make the MLM generate different values for our target distributions of tokens. We consider two types of mask patterns: the K-offset pattern and the Multimask pattern. 

\begin{enumerate}
    \item The K-offset mask pattern additionally masks the last \(K\) tokens from the input sequence (\(K = 3\) in the second row in Figure \ref{fig:conditionals}), and feed them to the MLM as \textbf{\textit{given}} output. For example, for Encoder-Decoder models like UL2, we feed \(K\) starting tokens to the decoder instead of the usual 0\footnote{For decoder-only MLMs like PaLM2, the input tokens and the \(K\) tokens are simply concatenated.}. The model then generates a different version of our distribution of interest. Because of the inconsistency issue, this new distribution is often remarkably different than the one from the baseline.
    \item The Multimask pattern additionally masks \(N\) random spans in the input sequence (\(N=1\) in the third row in Figure \ref{fig:conditionals}). This pattern is also parameterized by span length \(S\) and gap length \(G\) between spans. Recall that masking multiple spans is a common practice during the pretraining of MLMs. When our input contains multiple \texttt{[MASK]}'s, we feed additional tokens to the decoder, which correspond to the masked tokens in the \(N\) spans. The Multimask pattern will prompt the MLM to generate another different version of our conditional of interest.
\end{enumerate}

While our K-offset and Multimask conditionals may seem contrived at first glance, they potentially represent different knowledge learned by the model during pretraining. And as we will show in the sections next, they often contradict the baseline conditional and can be complementary to it.

To begin with, we select a number of specific mask patterns through parameterization of K-offset and Multimask patterns. This is done on the validation set of the evaluation dataset based on their individual accuracies. For any dataset in Lambada, MMLU and BigBench and any model in UL2 and T5, we always consider 10 types of conditional distributions as our set of interest. For example, for the combination of UL2 and Lambada we consider the baseline conditional, 6 K-offset conditionals (\(K \in[1,6]\)), and 3 Multimask conditionals (\((N, S, G)  \in \{(3, 5, 1), (3, 5, 2), (3, 10, 1)\}\)). See the full list of patterns in Appendix \ref{sec:mask_patterns}.

\subsection{Exposing inconsistencies}
\label{sec:exposing}

To quantitatively expose the severity of the inconsistencies among the numerous conditionals we use 3 benchmark datasets. 

\begin{enumerate}
    \item Lambada (LAnguage Modeling Broadened to Account for Discourse Aspects): Lambada is a dataset crafted to test the capabilities of computational models in language understanding, particularly in predicting the final word of a text passage when it requires understanding the broader context. This dataset focuses on the challenge of word prediction requiring a broad discourse context, aiming to evaluate if models can effectively utilize long-range dependencies in text. 
To evaluate inconsistency and ensembling on Lambada, we use the baseline conditional to generate on average 5 last words as candidates.

\item MMLU (Massive Multitask Language Understanding): The MMLU benchmark represents a leap towards evaluating the comprehensive knowledge acquired by models during pretraining. It encompasses a wide array of subjects, spanning elementary to advanced professional levels across STEM, humanities, and social sciences. MMLU aims to understand the depth and breadth of models' knowledge and reasoning abilities. MMLU is a multiple choice dataset from which the model chooses the best answer given the input question.

\item BigBench (Beyond the Imitation Game Benchmark): BigBench focuses on challenging tasks and aim to evaluate and understand models' performance across a spectrum of complexities and subject areas. This benchmark is designed not just to test models but also to highlight potential areas for future research and development. Similar to MMLU, BigBench is also a multiple choice dataset.

\end{enumerate}

We show that these incoherent conditionals often disagree on which answer is the best for a multi-choice question in MMLU and BigBench or which word is the best for last word prediction in Lambada. We demonstrate such incoherence on the three datasets by measuring how often the distributions cannot agree on the prediction. 

\begin{figure*}[h]
    \centering
    \includegraphics[width=1\linewidth]{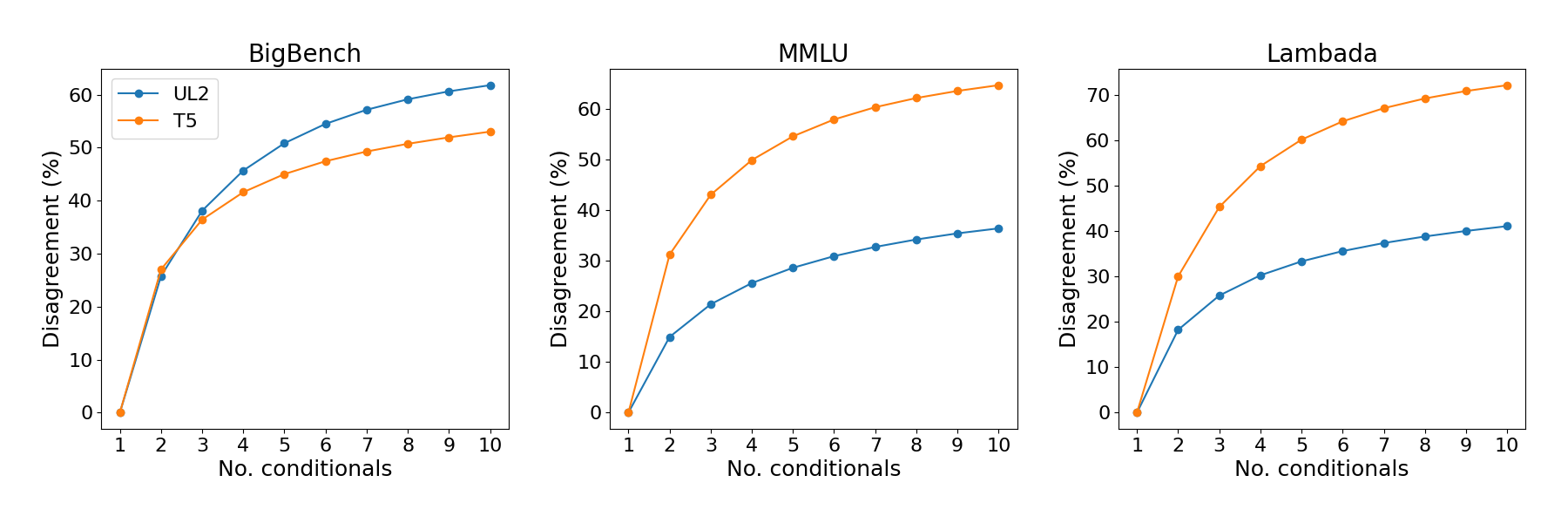}
    \caption{Different conditionals disagree on the prediction to make.}
    \label{fig:disagreement}
\end{figure*}

For example, consider a toy last word prediction task mimicking Lambada: \texttt{The cutest cat breed in the world is the [MASK]}. While the baseline conditional might rank \texttt{Munchkin} the highest, another conditional under our consideration might rank \texttt{Persian} the highest. We choose between 2 to 10 conditionals from our set of interest. When we choose less than 10 conditionals, all possible combinations are run and the results are aggregated. We count how often the conditionals cannot agree on the prediction. Figure \ref{fig:disagreement} shows that there exists considerable disagreement among the different conditionals we consider. And as expected, the more conditionals are considered, the more likely there is disagreement. But even with 2 conditionals, the disagreement can be as high as 20\%. However, disagreement converges to an upper bound. This means that there exists some questions in every benchmark on which the model is ``confident'' on its answer.

\subsection{Ensemble of Conditionals}l
\label{sec:ensemble}

Since we have shown that there are considerable inconsistencies among the conditionals corresponding to different masking patterns, it is worth investigating the potential benefit of ensembling them at inference time.

The gist of Ensemble of Conditionals (EOC) is to put the numerous raw conditionals provided by a trained MLM through an ensemble heuristic. EOC can be seen as a self-ensemble approach where the different outputs provided by one model are ensembled together, similar to ensembling outputs from multiple models in traditional ensemble learning.

To ensemble different conditionals for a final prediction, we use the max-pooling approach\footnote{Outperforms average-pooling in our pilot experiments.}. Consider that
the \(i\)th competing conditional assigns probability \(p_{ij}\) to the \(j\)th candidate completion (either a last word candidate in Lambada or an answer from a multiple choice question in MMLU). The winning conditional and final completion prediction is 
\begin{eqnarray}
 \hat{i}, \hat{j} = \argmax p_{ij},
\label{eq_argmax}
\end{eqnarray}

\begin{figure*}[h]
    \centering
    \includegraphics[width=1\linewidth]{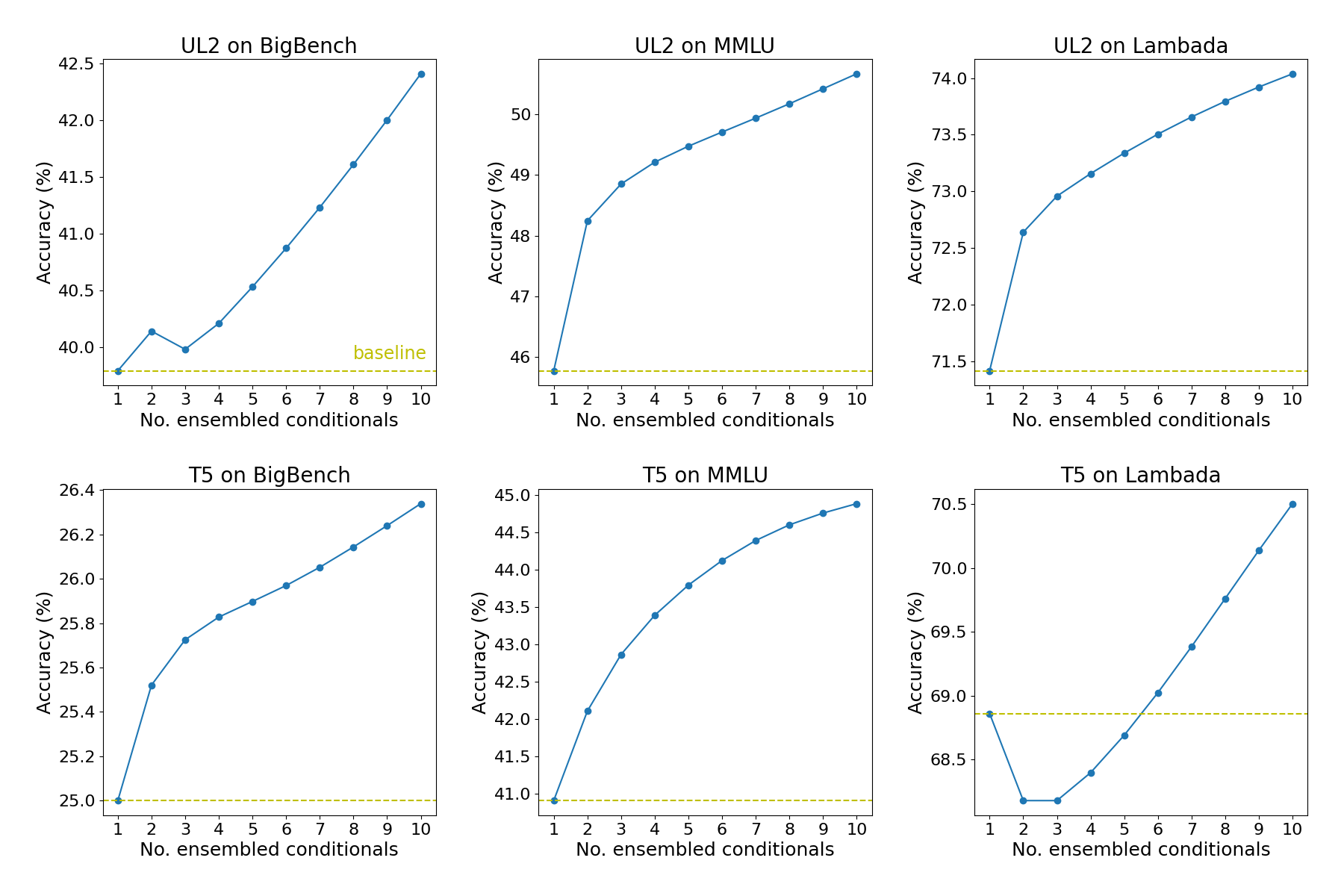}
    \caption{EOC improves MLM accuracy}
    \label{fig:eoc_accuracy}
\end{figure*}

In our experiments, we progressively ensemble more conditionals to observe accuracy changes. Similar to in the experiment on disagreement, when the number of ensembled conditionals is less that the total 10, all combinations are run and the results are aggregated. Results in Figure \ref{fig:eoc_accuracy} show that EOC can improve the accuracy of the model's final prediction. In additional, more ensembled conditionals can lead to higher accuracy.

\section{Inconsistencies in BERT-style MLMs} \label{sec:BERT}

\begin{figure*}
\centering
\includegraphics[width=1.0\textwidth]{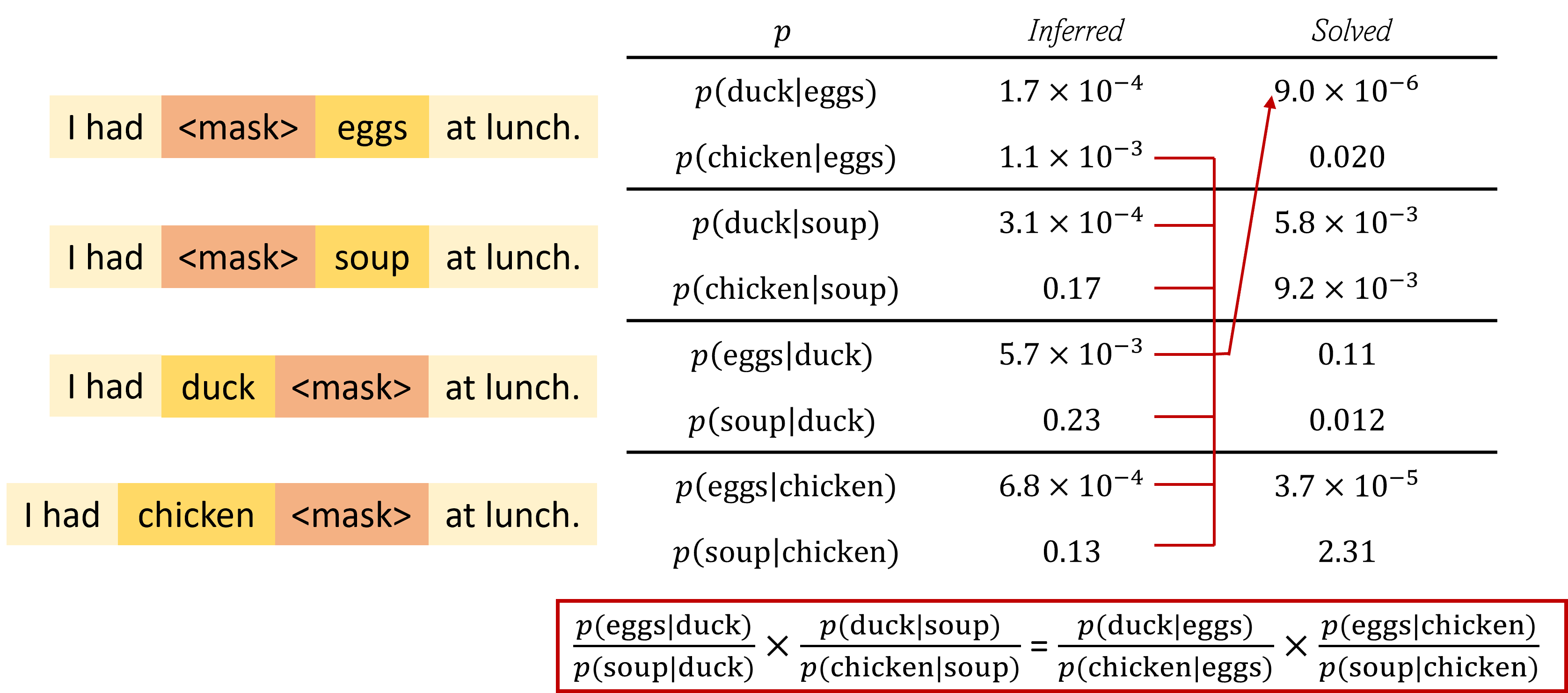}
\caption{Inconsistencies in the BERT-style MLM. Each ``inferred'' value refers to the probability given by the MLM (RoBERTa-large in this figure). Each ``solved'' value is obtained by passing the other 7 ``inferred'' values to the equation in the red square. We see that the difference between each inferred and solved value is significant (the solved value may even be larger than 1).}\label{fig:RoBERTa_example}
\end{figure*}

T5-style MLMs have the flexibility of generating sequences of variable length and are very useful in practice. Although researchers mainly focus on T5-style MLMs in the current era, we touch on inconsistencies in BERT-style MLMs in this section because of the historical impactfulness of the BERT model and their unique architecture with only bidirectional attention.

While BERT-style models can only model the distributions of individual tokens by their default design, there has been research effort \cite{goyal2021exposing, wang2019bert, yamakoshi2022probing} on sampling sequences from it by modeling its implicitly specified joint distribution one way or another. For example, \citet{goyal2021exposing} views it as an energy-based model defined using the bidirectional conditionals of the masked tokens. Such research effort is based on the intuition that bidirectional conditionals could be more robust than unidirectional conditionals \cite{goyal2021thesis}. This line of research has operated based on the assumption that the overly abundant bidirectional conditionals that the BERT-style MLMs provide are self-consistent. 

We demonstrate in this section that this is not the case at all. There are considerable inconsistencies that exist among the bidirectional conditionals that a trained BERT-style model provides. Figure \ref{fig:RoBERTa_example} demonstrates such an example. Since BERT-style models do not easily offer token distributions for completions, here we use bigrams in raw unstructured text to expose the inconsistencies instead of on standard benchmarks.

We consider 4 bigrams in a surrounding context: $x_{11}x_{21}$, $x_{11}x_{22}$, $x_{12}x_{21}$ and $x_{12}x_{22}$. $x_{11}$ and $x_{12}$ are two possible tokens that the first position can take; $x_{21}$ and $x_{22}$ the second. One can easily verify\footnote{Clue: converting each fraction term using the basic law in Figure \ref{fig:t5_example}. Equation \ref{eq5} was discussed in \cite{arnold1989compatible}.} that the 8 conditional distributions concerning such four bigrams should theoretically satisfy
\begin{align}\label{eq5}
\begin{aligned}
    \dfrac{p(x_{21}|x_{11})}{p(x_{22}|x_{11})} \times \dfrac{p(x_{11}|x_{22})}{p(x_{12}|x_{22})}  = \\ 
    \dfrac{p(x_{11}|x_{21})}{p(x_{12}|x_{21})} \times \dfrac{p(x_{21}|x_{12})}{p(x_{22}|x_{12})}
\end{aligned}
\end{align}

\begin{table*}[h]
\centering
\caption{Difference of log-probabilities between inferred and solved conditionals. The difference would be 0 for self-consistent MLMs. Roughly a 0.8 difference means that one is 120\% larger than the other.}
\begin{tabular}{ccc}
\hline
\textbf{Metric} &  \textbf{RoBERTa-base} & \textbf{RoBERTa-large} \\ \hline 
log-probability difference ($d_{\log p}$)       & 0.836 & 0.792 \\ \hline
\end{tabular}
\label{tab:RoBERTa_statistics}
\end{table*}

One way to test the inconsistencies among the 8 conditionals is to try to solve one using the other 7 and compare the solved conditional with the original (inferred by model) one. We show the solved conditionals in the example in Figure \ref{fig:RoBERTa_example}.
It clearly demonstrates that the probabilities given by a BERT-style MLM can be in serious inconsistencies with each other. 

We use the first segment of the validation partition of the C4 \cite{raffel2020exploring} dataset as the unstructured text corpus for quantification. Our goal here is to come up with quadruples of bigrams in the form of ($x_{11}x_{21}$, $x_{11}x_{22}$, $x_{12}x_{21}$, $x_{12}x_{22}$) in a certain context. We perform a full bigram sweep for the sequence. We always include the original bigram into the quadruple. To find the 3 alternative bigrams, we mask the whole original bigram, and generate alternatives using BART \cite{lewis2019bart}. In practice, we use beam search in \texttt{bart.generate()} with beam size 50. Note that BART by default is a T5-style MLM therefore it can generates multiple tokens for one mask. We keep all resulting generations that are 2 tokens.  We verify if there is a quadruple of bigrams in the generations in the said fashion and add them to our diagnostics dataset if so. We end up with 7431 quadruples. The following is an example in our diagnostics dataset.

\textit{Original sequence}: \texttt{Brown cats are the most common type of pets in America.}

\textit{Original bigram}: \texttt{Brown cats ($x_{11}x_{21}$).}

\textit{Alternative bigrams}: \texttt{Brown dogs ($x_{11}x_{22}$), White cats ($x_{12}x_{21}$),  White dogs ($x_{12}x_{22}$).}

To obtain conditionals in Figure \ref{fig:RoBERTa_example}, we mask the bigram with two \texttt{[MASK]}'s and feed the sequence to Roberta.

Note that there are many variables that go into building the diagnostics dataset. Our approaches were automatic but also somewhat unprincipled. We are not surprised if variations in the diagnostics dataset could result in some differences in the evaluation results. 

We quantify inconsistencies using difference of log probabilities\footnote{Here using logarithm makes it robust against changes in scale. One may also use other metrics for quantification.}. 
\begin{align}\label{eq6}
\begin{aligned}
    d_{\log p} = |\log p_{\texttt{solved}} - \log p_{\texttt{inferred}}|
\end{aligned}
\end{align}

Table \ref{tab:RoBERTa_statistics} shows the results, which clearly indicate strong inconsistencies among the bidirectional conditionals provided by the RoBERTa model.

\section{Summary \& Discussions}

This paper focused on the inconsistency problem concerning the conditionals provided by MLMs. We demonstrated and quantified the inconsistencies that exist in large MLMs. Based on our observations, we propose an inference-time approach that ensembles multiple inconsistent conditionals to improve the models' performance. The inconsistencies originate from the fact that the number of bidirectional conditionals MLMs can learn far exceeds what is needed for constructing the joint distribution. Given the recent evidence that MLM-based pretraining is a useful paradigm, we think that resolving its inconsistency issue could be a necessary step for future work. While our inference-time ensembling approach improves accuracy, it can only be seen as a limited patch-up method that only unite a certain number of selected conditionals. We believe that for long-term research, this problem should be ideally addressed as part of the expensive pretraining stage, for which our experiment techniques and results can be seen as a reference.

Such inconsistencies may remind readers of GPT's sensitivity to prompts \cite{openai2023gpt4}. It's crucial to understand that those sensitivities refer to inconsistencies in the space of semantics, which are distinct from the focus of our discussion. The inconsistencies highlighted in this paper address the peculiarities of MLMs in the fundamental space of token distributions.



\section*{Limitations}

\begin{enumerate}
    \item The discussion in Section \ref{sec:why} only specified a prerequisite for inconsistencies. As for why such inconsistencies mechanistically form during training and how they might be mitigated or avoided during training, we leave the research to future work.
    
    \item Although we tested mid-sized MLMs such as UL2-20B, it is no secret that some powerful masked language models like U-PaLM and PaLM2 are kept out of open access and they might behave somewhat differently. We leave diagnostics on those models for researchers with access. We don't expect the issue to completely disappear for those models.

    \item Apart from pretraining, it has been shown that paradigms like instruction tuning \cite{wei2021finetuned} and reinforcing \cite{ouyang2022training} can improve the performance of language models. How those techniques interplay with the inconsistency phenomenon is worth looking into.

\end{enumerate}


\section*{Impact Statements}

This paper presents work whose goal is to advance the field of Machine Learning. There are many potential societal consequences of our work, none which we feel must be specifically highlighted here.

\nocite{langley00}

\bibliography{custom}

\newpage
\appendix

\onecolumn

\section{Why not use Llama in the experiments?}
Llama and Llama2 are causal autoregressive LLMs that did not utilize the MLM training objective (not mentioned in paper). We expect the MLM pretraining objective to be a useful supplementary to them. 
\section{No. bidirectional conditionals specified by MLMs}
\label{sec:no_conditionals}

$N_{mlm}(1)$ is given by:
\begin{align}
N_{mlm}(1)= {L} \times |V|^{L-1} \times (|V| - 1) \nonumber \\ 
= {L} \times (|V|^L - |V|^{L-1})
\label{eq:Nmlm1_app}
\end{align}

$L$ represents how many positions the predicted one token could be in. The number of variations of the surrounding context of length $L-1$ is $|V|^{L-1}$. Given the surrounding context and the position of the predicted token, the number of free conditionals is $|V|-1$ (we assume a BERT-style MLM here; a T5-style MLM naturally provides distributions of tokens of a variable amount). Multiplying the 3 numbers together gives Equation \ref{eq:Nmlm1_app}.

One may also consider $N_{mlm}(k)$ for BERT-style MLMs, where the $k$ masked tokens can be anywhere in a sequence of $L$ tokens. Note that BERT-style MLMs by default do not model the joint distribution of the $k$ tokens. Instead it models their individual marginal distributions conditioned on the context, which we let $N_{mlm}(k)$ denote the number of here.

$N_{mlm}(k)$ is given by:
\begin{align}
N_{mlm}(k)= {L \choose k} \times |V|^{L-k} \times (|V| - 1)^{k}  
\label{eq:Nmlmk_app}
\end{align}

In Equation \ref{eq:Nmlmk_app} (same as Equation \ref{eq2}), ${L \choose k}$ represents how many combinations of positions the predicted $k$ tokens could be in. The number of variations of the surrounding context of length $L-k$ is $|V|^{L-k}$. Given the surrounding context and the positions of the predicted tokens, the number of free conditionals is $(|V| - 1)^{k}$.

One can easily see that the number of conditionals an MLM provides far exceeds what is needed for defining a joint distribution, which sets up room for there to be inconsistencies among them. We omit detailed discussions for the number of conditionals provided by T5-style MLMs here.

\section{Mask patterns}
\label{sec:mask_patterns}

\begin{enumerate}
    \item UL2 on MMLU.
            \(K \in[1,6]\);
           \((N, S, G)  \in \{(3, 5, 1), (3, 5, 2), (3, 10, 1)\}\)
    \item UL2 on Lambada.
            \(K \in[1,6]\);
           \((N, S, G)  \in \{(3, 5, 1), (3, 5, 2), (3, 10, 1)\}\)
    \item UL2 on BigBench.
\(K \in[1,3]\);
\((N, S, G)  \in \{(3, 5, 1),(3, 5, 2),(3, 3, 1),(3, 3, 2),(3, 4, 1),(3, 4, 2)\}\)

    \item      T5 on MMLU.
\(K \in[1,3]\);
\((N, S, G)  \in \{(3, 5, 1),(3, 5, 2),(3, 3, 1),(3, 3, 2),(3, 4, 1),(3, 4, 2)\}\)

    \item T5 on Lambada.
             \(K \in[1,6]\);
           \((N, S, G)  \in \{(3, 5, 1), (3, 5, 2), (3, 10, 1)\}\)

    \item T5 on BigBench.
\(K \in[1,3]\);
\((N, S, G)  \in \{(3, 5, 1),(3, 5, 2),(3, 3, 1),(3, 3, 2),(3, 4, 1),(3, 4, 2)\}\)

Some subjects (subsets) in MMLU and BigBench are very challenging for mid-sized models like UL2-20B and T5-13B. We report on subjects that the baseline has a decent performance on (accuracy > 0.4).

\end{enumerate}

\end{document}